\documentclass[journal]{IEEEtran}
\usepackage{graphicx}

\usepackage{color}
\usepackage[table]{xcolor}
\usepackage{colortbl}
\usepackage{bm}
\usepackage{amsmath}
\usepackage{epstopdf}

\usepackage{multirow}
\usepackage{booktabs}
\usepackage{amssymb}

\usepackage{bbding}
\usepackage{stfloats}

\usepackage{lineno,hyperref}

\ifCLASSINFOpdf
\else
\fi
\begin{document}
%
\title{Multiple-Exit Tuning: Towards Inference-Efficient Adaptation for Vision Transformer}
%
%
%

\author{Zheng~Liu,
        Jinchao Zhu,
        Nannan Li,
        and~Gao~Huang,~\IEEEmembership{Member,~IEEE}
\thanks{
Z. Liu, J. Zhu, N. Li, and G. Huang are with the Department of Automation, BNRist, Tsinghua University, Beijing 100084,
China. G. Huang is also with Beijing Academy of Artificial Intelligence.
}
\thanks{Corresponding author: G.~Huang (e-mail: gaohuang@tsinghua.edu.cn).}
}

%
%

\markboth{Journal of \LaTeX\ Class Files,~Vol.~xx, No.~xx, September~2024}%
{Shell \MakeLowercase{\textit{et al.}}: Bare Demo of IEEEtran.cls for IEEE Journals}
%



\maketitle

\begin{abstract}

Parameter-efficient transfer learning (PETL) has shown great potential in adapting a vision transformer (ViT) pre-trained on large-scale datasets to various downstream tasks. Existing studies primarily focus on minimizing the number of learnable parameters. Although these methods are storage-efficient, they allocate excessive computational resources to easy samples, leading to inefficient inference. To address this issue, we introduce an inference-efficient tuning method termed multiple-exit tuning (MET). MET integrates multiple exits into the pre-trained ViT backbone. Since the predictions in ViT are made by a linear classifier, each exit is equipped with a linear prediction head. In inference stage, easy samples will exit at early exits and only hard enough samples will flow to the last exit, thus saving the computational cost for easy samples. MET consists of exit-specific adapters (E-adapters) and graph regularization. E-adapters are designed to extract suitable representations for different exits. To ensure parameter efficiency, all E-adapters share the same down-projection and up-projection matrices. As the performances of linear classifiers are influenced by the relationship among samples, we employ graph regularization to improve the representations fed into the classifiers at early exits. Finally, we conduct extensive experiments to verify the performance of MET. Experimental results show that MET has an obvious advantage over the state-of-the-art methods in terms of both accuracy and inference efficiency.

\end{abstract}


%
\IEEEpeerreviewmaketitle

\section{Introduction}
%
%
%
%
\IEEEPARstart{V}{ision} transformer (ViT) \cite{b1,b2} has achieved remarkable success in various recognition tasks \cite{b3,b4,b5} due to its excellent modelling abilities. Unfortunately, it is hard for us to train a well-performing ViT from scratch on the datasets with limited samples \cite{b6}. To solve this dilemma, a common practice is to pre-train a ViT on large-scale datasets like ImageNet \cite{b7} and then fully fine-tune the pre-trained model for downstream domains. However, this practice is storage-inefficient because we need to store the whole tuned model for each downstream task.

Parameter-efficient transfer learning (PETL) \cite{b8,b9,b10} aims to adapt large pre-trained models to downstream tasks using only a fraction of trainable parameters. It has been the predominant method for transferring the knowledge contained in pre-trained ViTs \cite{b11,b12}. Compared to full fine-tuning, PETL is much more storage-friendly since we just need to store the learnable parameters. Most existing PETL methods focus on reducing the number of trainable parameters as much as possible. They either insert extra learnable blocks \cite{b13,b14,b15} or tokens \cite{b16,b17} into a frozen ViT backbone or selectively tune some modules of a pre-trained ViT while freezing the others \cite{b18}. At test stage, these methods will allocate the same computational resources to all inputs. A pre-trained ViT can be viewed as a feature extractor and our goal is to extract suitable representations from it \cite{b6}. The representation ability of a ViT is supported by its encoder layers. This begs the question: is it necessary to represent all inputs with all encoder layers? In real-world applications, a large number of samples are easy to recognize \cite{b19,b20}, which means that their representations can be easily learned. Ideally, if all inputs are represented with suitable encoder layers, we will save lots of computational resources on easy samples and thus boost the inference efficiency of the fine-tuned model.

Motivated by early-exiting dynamic neural networks \cite{b21,b22}, we propose to allocate suitable encoder layers to inputs by incorporating multiple exits into a pre-trained ViT backbone. Easy samples will exit at early exits when they are represented properly while only sufficiently challenging samples will pass through all encoder layers. In this way, we can reduce the computational resources for easy samples and speed up the whole inference process. In ViT, the representations of inputs are extracted by class tokens and there is only one class token for each sample. For a neural network with multiple exits, it is optimized by minimizing the cross-entropy losses of all exits \cite{b21}. Nevertheless, the representation-learning processes of a sample at different exits may negatively influence each other due to the gradient conflicts caused by different losses, which will be discussed in Section IV.D.

\begin{figure}[!h]
\centering
\includegraphics[trim={3.5cm 7.0cm 1cm 3.5cm},clip,width=7.0in]{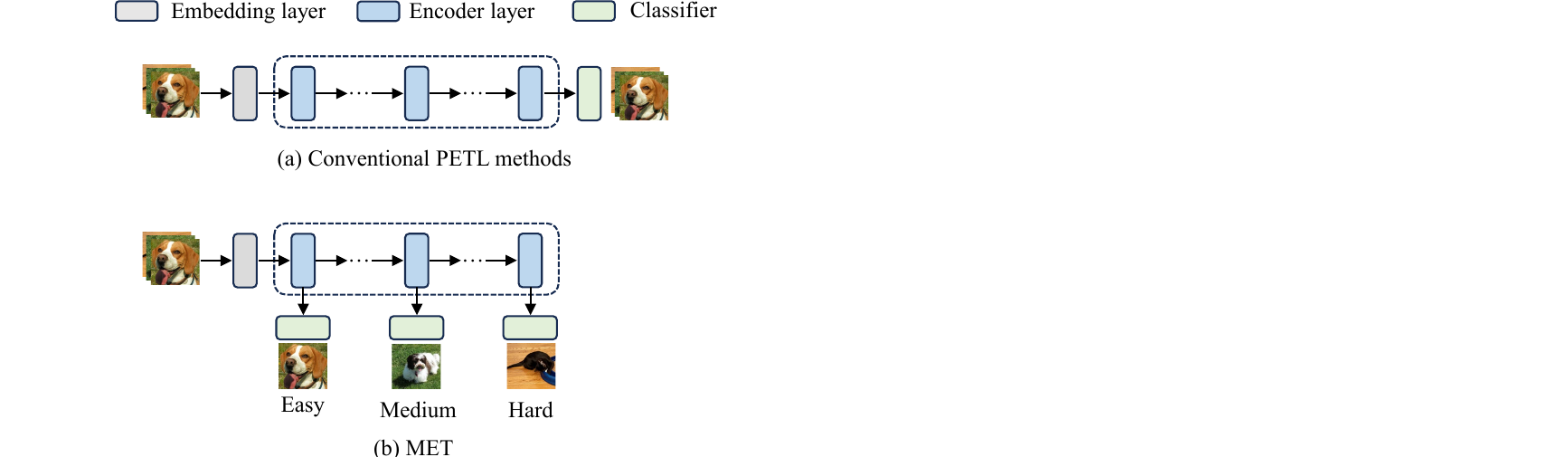}
\caption{Inference comparison between MET and conventional PETL methods.}
\label{fig1}
\end{figure}

In this paper, to address the aforementioned problem, we propose a simple yet effective method named multiple-exit tuning (MET), to transfer the representative knowledge contained in a pre-trained ViT to downstream tasks. Our proposed method is totally different from the existing ones. Fig.~\ref{fig1} shows the inference difference between MET and conventional PETL methods. In MET, multiple exits are inserted into the pre-trained backbone and samples flow to different exits depending on whether they are easy to recognize or not. Easy samples will flow to early exits and only hard enough samples will exit at the last exit. By contrast, all samples in conventional PETL methods will be recognized by the classifier after the last encoder layer.

MET includes exit-specific adapters (E-adapters) and graph regularization. All E-adapters share the same down-projection and up-projection matrices for the purpose of efficient storage. Graph regularization is applied to enhance the intra-class compactness and inter-class separability of data points at early exits. To sum up, the main contributions are as follows.

\begin{itemize}

 \item We introduce parameter-efficient E-adapters to disentangle the transformations of class token, facilitating the extraction of representations for each exit.

 \item To improve the learning abilities of early classifiers, two self-supervised graphs are constructed to regularize the exit-specific representations in training process.

 \item Extensive experiments conducted on 28 downstream tasks demonstrate that MET is more competitive than the state-of-the-art methods.
\end{itemize}

The rest of this paper is organized as follows. Related work and preliminaries are reviewed in Section II and Section III. The proposed method is introduced in Section IV. Experiments are conducted in Section V. Conclusion and future work are given in Section VI.

\section{Related work}

\subsection{Parameter-Efficient Transfer Learning}

Over the past several years, parameter-efficient transfer learning (PETL) has mainly focused on adapting the transformers pre-trained on large-scale datasets to various downstream domains with a few learnable parameters. The existing studies can be roughly categorized into four branches: partial tuning, prompt tuning, extra-module tuning, and mixed tuning methods.

\textbf{Partial tuning} is a simple type of PETL method which directly tunes a small subset of the parameters existing in backbones while freezing the others. The methods belonging to this category include LN-tune \cite{b18}, BitFit \cite{b23}, DiffFit \cite{b24}, linear probe \cite{b25}, etc. \textbf{Prompt tuning} introduces a set of trainable tokens to the inputs of transformer modules. For example, Jia et al. \cite{b17} tune a pre-trained backbone by prepending learnable tokens to the inputs of encoder layers; Han et al. \cite{b26} add key-value prompts and visual prompts to multi-head attention (MHA) modules and encoder layers, respectively; Tu et al. \cite{b27} prepend extra query tokens for each MHA operator to aggregate the intermediate representations of encoder layers; Li et al. \cite{b28} directly prepend continuous task-specific vectors to the input of the entire pre-trained backbone and only optimize these added vectors when tuning on the datasets of downstream tasks. By inserting extra blocks into a pre-trained backbone, \textbf{extra-module tuning} has attracted increasing attention from researchers. Adapter \cite{b29}, side-tuning \cite{b30} and reparameterization \cite{b31} are popular research topics in extra-module tuning. The input of adapter is first mapped into a low-dimensional space through a down-projection matrix followed by a nonlinear function, and then remapped into the original space by an up-projection matrix. Each transformer layer consists of an MHA block, a feed-forward network (FFN) module and two layer normalization operators. Adapter often adjusts the representations of samples from downstream tasks by tuning the inputs of encoder layers \cite{b32} or the outputs of MHA and FFN \cite{b29}, or by learning the complementary information of FFN \cite{b33} or transformer layers \cite{b34,b35}. In side-tuning, side networks, such as E3VA \cite{b36}, DTL \cite{b37}, SAN \cite{b38} and LSA \cite{b39}, are designed to extract the useful information existing in pre-trained backbones. Reparameterization introduces learnable blocks during training phase, but these modules are perfectly incorporated into the whole pre-trained model at inference stage, avoiding extra computational cost. In the related work about reparameterization, such as Hydra \cite{b40}, FacT \cite{b41}, RLRR \cite{b42}, LoRA \cite{b43}, the inserted blocks are proposed with multiple linear transformations. \textbf{Mixed tuning} refers to the methods which are the combination of different PETL methods. For instance, UniPELT \cite{b44} incorporates LoRA, prefix tokens and adapter; NOAH \cite{b45} adapts neural architecture search techniques to find the optimal combination of LoRA, adapter, and VPT \cite{b17}.

Most existing studies on PETL are proposed to solve the storage-expensive problem. In these studies, all inputs share the same computational resources in inference stage. In fact, easy samples can be correctly classified with less computational cost than hard samples, and thus these methods are inference-inefficient for easy samples. Dynamic tuning (DyT) \cite{b46} is recently introduced to reduce the redundant computation in inference process by activating or deactivating tokens dynamically. Our method (MET) is designed to assign suitable encoder layers for different samples. DyT and MET are fundamentally different.

\subsection{Early-Exiting Dynamic Neural Networks}

Dynamic neural networks can achieve a good trade-off between inference efficiency and learning performance by adapting their structures or parameters for different inputs during inference stage \cite{b47}. As a typical dynamic neural network, early-exiting dynamic neural network (EDNN) \cite{b48,b49,b50} inserts multiple exits along the depth direction. In training phase, all losses of exits are jointly optimized. During inference, if a sample's prediction confidence obtained by the classifier assigned at a certain exit exceeds a threshold, this sample will exit at this exit; otherwise, it will flow to the next exit. For more details, please refer to \cite{b50}.

Based on the above schemes, easy samples will exit at early exits and only sufficiently difficult samples will flow to the last exit, thereby saving the computational cost for easy samples. In this paper, by incorporating the learning schemes of EDNN into transfer learning, we introduce a storage-friendly and inference-efficient tuning method.

\section{Preliminaries}
\subsection{Vision Transformer}

A vanilla vision transformer (ViT) \cite{b2} consists of a patch embedding (PE) layer, encoder layers and a prediction head. As the basic layer in ViT, PE layer transforms an input image ${\bf{X}} \in {\mathbb R^{h \times w \times 3}}$ into a sequence of $d$-dimensional tokens  $[{{\bf{x}}_1}{\bf{P}}; \cdots ;{{\bf{x}}_n}{\bf{P}}] \in {\mathbb R^{n \times d}}$, where $(h,w)$ is the resolution of ${\bf{X}}$,  ${{\bf{x}}_i} \in {\mathbb R^{1 \times 3{m^2}}}$ is the $i$-th flattened patch, $(m,m)$  is the resolution of each image patch, ${\bf{P}} \in {\mathbb R^{3{m^2} \times d}}$ is the projecting matrix and $n = \frac{{hw}}{{{m^2}}}$. The mathematical expression for this layer is as follows.
\begin{equation}
[{{\bf{c}}_{PE}};{{\bf{Z}}_{PE}}] = [{\bf{c}};{{\bf{x}}_1}{\bf{P}}; \cdots ;{{\bf{x}}_n}{\bf{P}}] + {{\bf{E}}_{pos}}, \label{eq1}
\end{equation}
where ${\bf{c}} \in {\mathbb R^{1 \times d}}$ is the prepended learnable class token; ${{\bf{E}}_{pos}} \in {\mathbb R^{(n + 1) \times d}}$ represents the position embeddings; ${{\bf{c}}_{PE}}$ and   ${{\bf{Z}}_{PE}}$ are the transformed class token and features in PE layer.

Layer norm (LN), MHA and FFN are three basic components for each encoder layer. In the $k$-th encoder layer, we have
\begin{equation}
[{{{\bf{\hat c}}}_k};{{{\bf{\hat Z}}}_k}] = {\rm{MHA}}({\rm{LN}}([{{\bf{c}}_{k - 1}};{{\bf{Z}}_{k - 1}}])) + [{{\bf{c}}_{k - 1}};{{\bf{Z}}_{k - 1}}],
\label{eq2}
\end{equation}
\begin{equation}
[{{\bf{c}}_k};{{\bf{Z}}_k}] = {\rm{FFN}}({\rm{LN}}([{{{\bf{\hat c}}}_k};{{{\bf{\hat Z}}}_k}])) + [{{{\bf{\hat c}}}_k};{{{\bf{\hat Z}}}_k}],
\label{eq3}
\end{equation}
where ${{\bf{\hat c}}_k}$ and ${{\bf{\hat Z}}_k}$ are the transformations about class token and features in MHA block; ${{\bf{c}}_k}$ and ${{\bf{Z}}_k}$ are the corresponding representations obtained by this layer; $1 \le k \le L$, $L$ is the total number of encoder layers; ${{\bf{Z}}_0} = {{\bf{Z}}_{PE}}$ and ${{\bf{c}}_0} = {{\bf{c}}_{PE}}$.

We have $[{{\bf{c}}_L};{{\bf{Z}}_L}] = {\rm{FFN}}({\rm{LN}}([{{\bf{\hat c}}_L};{{\bf{\hat Z}}_L}])) + [{{\bf{\hat c}}_L};{{\bf{\hat Z}}_L}]$ after $L$ encoder layers. In ViT, ${{\bf{c}}_L}$ is taken as the final representation to make predictions. The predicting result ${\bf{y}}$  can be obtained by
\begin{equation}
{\bf{y}} = {\rm{Head}}(\rm{LN}({{\bf{c}}_L})),
\label{eq4}
\end{equation}
where ${\rm{Head}}( \cdot )$  is  a linear classifier.

\subsection{Adapter}

\begin{figure}[!h]
\centering
\includegraphics[trim={8cm 5.5cm 2cm 2cm},clip,width=5.0in]{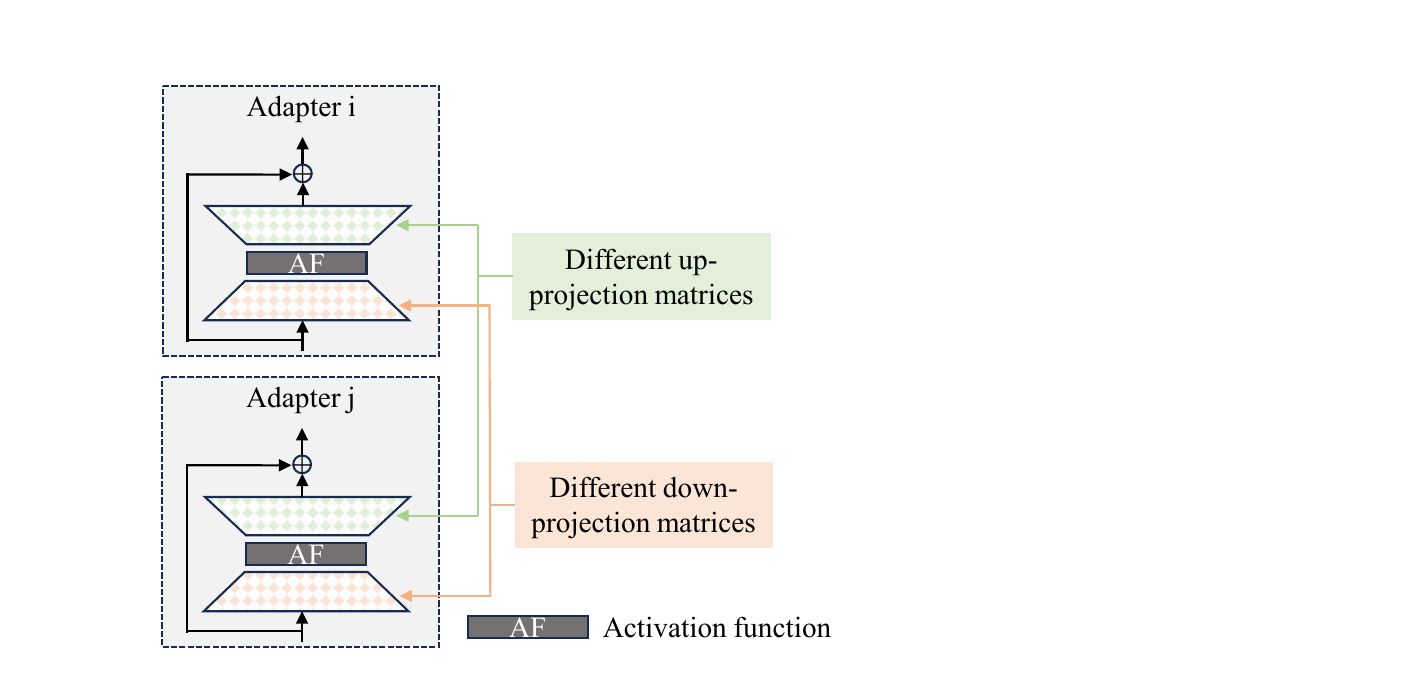}
\caption{The relationship between every two adapters.}
\label{fig2}
\end{figure}

Adapter is widely used to tune the intermediate representations contained in pre-trained backbones by two learnable projection matrices and one nonlinear activation function. Let ${{\bf{A}}_{input}} \in {\mathbb R^{(n + 1) \times d}}$ and ${{\bf{A}}_{ouput}} \in {\mathbb R^{(n + 1) \times d}}$ denote the input and output of an adapter, we have
\begin{equation}
{{\bf{A}}_{ouput}} = \sigma ({{\bf{A}}_{input}}{{\bf{M}}_{down}}){{\bf{M}}_{up}} + {{\bf{A}}_{input}},
\label{eq5}
\end{equation}
where $\sigma ( \cdot )$ is a nonlinear activation function,  ${{\bf{M}}_{down}} \in {\mathbb R^{d \times d'}}(d' \ll d)$ and ${{\bf{M}}_{up}} \in {\mathbb R^{d' \times d}}$ are down-projection and up-projection matrices. Inserting multiple adapters into a pre-trained backbone at different locations has become a popular strategy in PETL, where all projection matrices are different and trainable \cite{b29}. The relationship between every two adapters is depicted in Fig.~\ref{fig2}.

\section{Multiple-exit Tuning}

\begin{figure*}[!htb]
\centering
\includegraphics[trim={2.2cm 0.cm -0.1cm 0.1cm},clip,width=7.8in]{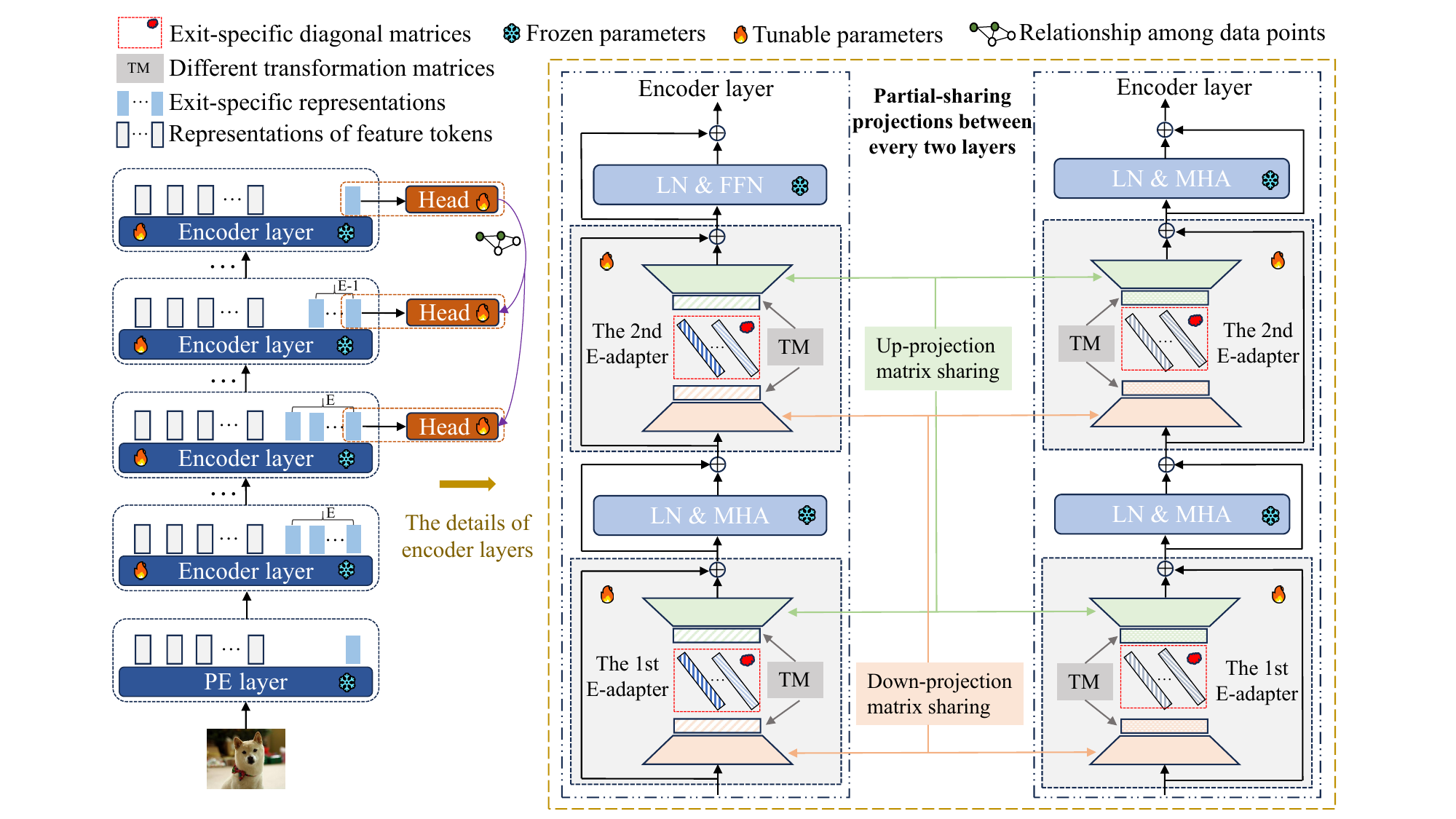}
\caption{The illustration of our proposed MET method.}
\label{fig3}
\end{figure*}

Our proposed method is shown in Fig.~\ref{fig3}. Multiple-exit tuning (MET) consists of two modules: exit-specific adapters (E-adapters) and graph regularization. The inserted E-adapters are used to extract discriminative representations for the classifiers at different exits. Since different exits require different representations about class token, the number of disentangled representations is equal to that of exits. Suppose there are $E$ exits in a pre-trained ViT backbone. All representations about class token and feature tokens pass through the same encoder layers together before reaching an exit. When arriving at an exit, the specific representation of class token is adopted to make predictions, while the others continue the forward propagation process. The graph regularization is introduced to boost the learning abilities of early classifiers.

\subsection{Exit-Specific Adapters}

Following the existing work on early-exiting dynamic neural networks \cite{b48,b49,b50}, when fine-tuning a pre-trained ViT with $E$ exits, we can define its objective function as
\begin{equation}
{\rm{O}} = \sum\limits_{e = 1}^E {\sum\limits_{j = 1}^N {\frac{1}{N}{\rm{C}}{{\rm{E}}_e}({\rm{Hea}}{{\rm{d}}_e}({\rm{LN}}({g_e}({{\bf{X}}_j},{{\bm{\theta }}_e}))),{y_j})} } ,
\label{eq6}
\end{equation}
where ${\rm{Hea}}{{\rm{d}}_e}( \cdot )$ is the $e$-th prediction head; ${\rm{C}}{{\rm{E}}_e}( \cdot )$ represents the cross-entropy loss of ${\rm{Hea}}{{\rm{d}}_e}( \cdot )$; ${g_e}( \cdot ,{{\bm{\theta }}_e})$ is the encoding function for the $e$-th exit with parameter ${{\bm{\theta }}_e}$; $N$ is the number of training samples; ${{\bf{X}}_j}$ represents the $j$-th training sample and ${y_j}$ is the ground-truth label of ${{\bf{X}}_j}$.

According to the learning scheme (see equation (4)) of ViT,  ${g_e}({{\bf{X}}_j},{{\bm{\theta }}_e})$ should be the representation of ${{\bf{X}}_j}$'s class token  for the $e$-th classifier. However, there is only one class token in ViT. If all classifiers share the representations of class token, their performances may negatively interact with each other, which will be discussed in Section IV.D.

To avoid the optimization conflicts caused by the shared representations, we introduce E-adapters to disentangle the transformations of this token, based on the work on adapter. There are two E-adapters inserted before MHA and MLP modules in each encoder layer. In each of these E-adapters, we expect to obtain exit-specific representations by different projection matrices. However, if we follow the relationship demonstrated in Fig.~\ref{fig2}, it will introduce a large quantity of trainable parameters, since $2L$ E-adapters are inserted into a pre-trained ViT backbone.

\begin{figure*}[!htb]
\centering
\includegraphics[trim={1.8cm 7cm 0cm 0.1cm},clip,width=7.6in]{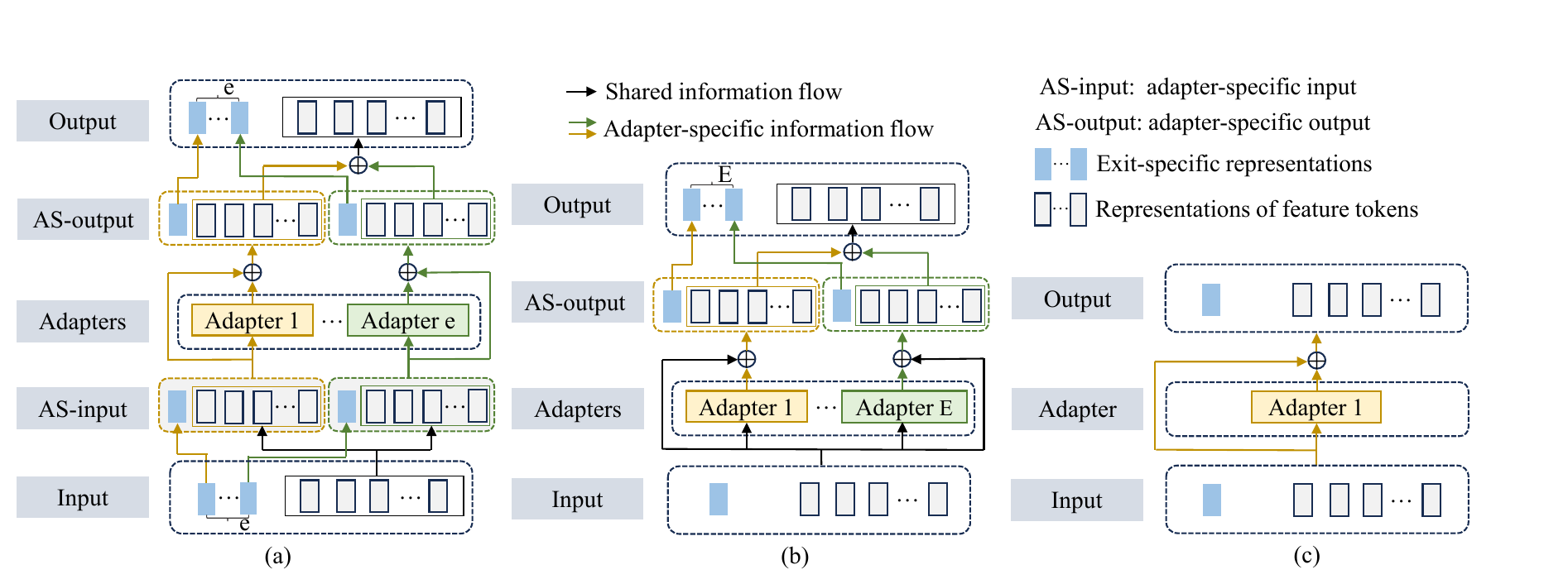}
\caption{The learning strategies of E-adapters: (a) the E-adapter containing $e$ ($1 < e < E$) exit-specific representations, (b) the first E-adapter at the first encoder layer, and (c) the second E-adapter at the last encoder layer. }
\label{fig4}
\end{figure*}

To reduce the number of learnable parameters, we design a partial-sharing projection scheme. As illustrated in Fig.~\ref{fig3}, all E-adapters share the same down-projection and up-projection matrices. Low-dimensional transformation matrices are applied to capture the differences among E-adapters. Diagonal matrices are used to adjust the suitable embeddings for different exits. Each E-adapter can be viewed as the combination of multiple adapters with shared projection matrices. As shown in Fig.~\ref{fig4}, the learning scheme of E-adapter is influenced by two factors: location and the number of exit-specific representations in input. If the input to the $m$-th E-adapter contains $e$ ($e > 1$) exit-specific representations (see Fig.~\ref{fig4}(a)), according to the property of adapter, we have
\begin{equation}
[{\bf{c}}_{out}^{m,i};{\bf{Z}}_{out}^{m,i}] = \sigma ({{\bf{Q}}_m}{{\bf{U}}^{down}}{{\bf{R}}_m}){{\bf{\Lambda }}_{m,i}}{{\bf{W}}_m}{{\bf{U}}^{up}} + {{\bf{Q}}_m},
\label{eq7}
\end{equation}
\begin{equation}
{\bf{C}}_{out}^m = [{\bf{c}}_{out}^{m,E - e + 1};{\bf{c}}_{out}^{m,E - e + 2}; \cdots ;{\bf{c}}_{out}^{m,E}],
\label{eq8}
\end{equation}
\begin{equation}
{\bf{Z}}_{out}^m = {\bf{Z}}_{out}^{m,E - e + 1} + {\bf{Z}}_{out}^{m,E - e + 2} +  \cdots  + {\bf{Z}}_{out}^{m,E},
\label{eq9}
\end{equation}
where ${{\bf{Q}}_m} = [{\bf{c}}_{in}^{m,i};{\bf{Z}}_{in}^m]$; ${{\bf{U}}^{down}} \in {\mathbb R^{d \times d'}}(d' \ll d)$ and  ${{\bf{U}}^{up}} \in {\mathbb R^{d' \times d}}$ are the shared down-projection and up-projection matrices; $[{\bf{c}}_{out}^{m,i};{\bf{Z}}_{out}^{m,i}]$ is the output of the ($i + e - E$)-th adapter and $E - e + 1 \le i \le E$; ${{\bf{R}}_m} \in {\mathbb R^{d' \times d'}}$ and ${{\bf{W}}_m} \in {\mathbb R^{d' \times d'}}$ are low-dimensional transformation matrices for this E-adapter; $\{ {{\bf{\Lambda }}_{m,i}}\} _{i = E - e + 1}^E$ are exit-specific diagonal matrices; ${\bf{c}}_{in}^{m,i}$ denotes the representation for the $i$-th exit and ${\bf{c}}_{out}^{m,i}$ is the corresponding transformation of ${\bf{c}}_{in}^{m,i}$; ${\bf{Z}}_{in}^m$ represents the input features; ${\bf{Z}}_{out}^{m,i}$ is the transformed features of the ($i-E+e$)-th adapter; ${\bf{C}}_{out}^m$ and ${\bf{Z}}_{out}^m$ are the final outputs. The first E-adapter at the first encoder layer is applied to obtain specific representations for exits (see Fig.~\ref{fig4}(b)). This E-adapter is formulated as follows
\begin{equation}
[{\bf{c}}_{out}^{1,i};{\bf{Z}}_{out}^{1,i}] = \sigma ({\bf{N}}{{\bf{U}}^{down}}{{\bf{R}}_1}){{\bf{\Lambda }}_{1,i}}{{\bf{W}}_1}{{\bf{U}}^{up}} + {\bf{N}},
\label{eq10}
\end{equation}
\begin{equation}
{\bf{C}}_{out}^1 = [{\bf{c}}_{out}^{1,1};{\bf{c}}_{out}^{1,2}; \cdots ;{\bf{c}}_{out}^{1,E}],
\label{eq11}
\end{equation}
\begin{equation}
{\bf{Z}}_{out}^1 = {\bf{Z}}_{out}^{1,1} + {\bf{Z}}_{out}^{1,2} +  \cdots  + {\bf{Z}}_{out}^{1,E},
\label{eq12}
\end{equation}
where ${\bf{N}} = [{{\bf{c}}_{PE}};{{\bf{Z}}_{PE}}]$ and $1 \le i \le E$. As for the second E-adapter at the last encoder layer (see Fig.~\ref{fig4}(c)), it is equivalent to one adapter, which is defined as
\begin{equation}
[{\bf{C}}_{out}^{2L};{\bf{Z}}_{out}^{2L}] = \sigma ({\bf{V}}{{\bf{U}}^{down}}{{\bf{R}}_{2L}}){{\bf{\Lambda }}_{2L,E}}{{\bf{W}}_{2L}}{{\bf{U}}^{up}} + {\bf{V}},
\label{eq13}
\end{equation}
where ${\bf{V}} = [{\bf{c}}_{in}^{2L,E};{\bf{Z}}_{in}^{2L}]$.

Compared to the vanilla adapters \cite{b29}, our proposed partial-sharing projection scheme is much more parameter-efficient. Suppose three exits are inserted into the last three encoder layers of a pre-trained ViT backbone. If we directly follow the relationship shown in Fig.~\ref{fig2}, we will have $12dd'(L - 1)$ extra parameters. There are $2dd' + 4Ld'd' + 6Ld' - 6d'$ learnable parameters in our designed scheme. Since $d' \ll d$, we have $2dd' + 4Ld'd' + 6Ld' - 6d' \ll 12dd'(L - 1)$. For example, when fine-tuning a pre-trained ViT-B/16 backbone \cite{b2}, the number of trainable parameters is reduced by approximately 98.48\%, demonstrating the powerful parameter-compression capability of the designed scheme.

For simplicity, we denote the mathematical formulation of the $m$-th E-adapter (containing $e$ exit-specific representations) as $[{\bf{C}}_{out}^m;{\bf{Z}}_{out}^m] = {{\rm{E}}_m}([{\bf{C}}_{in}^m;{\bf{Z}}_{in}^m])$ where ${\bf{C}}_{in}^m = [{\bf{c}}_{in}^{m,E - e + 1}; \cdots ;{\bf{c}}_{in}^{m,E}]$ and $1 \le e \le E$. In the $k$-th encoder layer of our proposed method, equations (2) and (3) become
\begin{equation}
{{\bf{J}}_k} = {\rm{MHA}}({\rm{LN}}({{\rm{E}}_{2k - 1}}({{\bf{F}}_k}))) + {{\rm{E}}_{2k - 1}}([{{{\bf{\bar C}}}_{k - 1}};{{\bf{Z}}_{k - 1}}]),
\label{eq14}
\end{equation}
\begin{equation}
[{{\bf{C}}_k};{{\bf{Z}}_k}] = {\rm{FFN}}({\rm{LN}}({{\rm{E}}_{2k}}({{\bf{J}}_k}))) + {{\rm{E}}_{2k}}({{\bf{J}}_k}),
\label{eq15}
\end{equation}
where ${{\bf{F}}_k} = [{{\bf{C}}_k};{{\bf{Z}}_k}]$ ($[{{\bf{C}}_k};{{\bf{Z}}_k}] = [{\bf{C}}_{in}^{2k - 1};{\bf{Z}}_{in}^{2k - 1}]$); ${{\bf{J}}_k} = [{{{\bf{\hat C}}}_k};{{{\bf{\hat Z}}}_k}]$ ($[{{{\bf{\hat C}}}_k};{{{\bf{\hat Z}}}_k}] = [{\bf{C}}_{in}^{2k};{\bf{Z}}_{in}^{2k}]$); ${{\bf{C}}_k} = [{\bf{c}}_k^e;{\bf{c}}_k^{e + 1}; \cdots ;{\bf{c}}_k^E]$  is the set of exit-specific representations in this encoder layer;  ${{\bf{\hat C}}_k} = [{\bf{\hat c}}_k^e;{\bf{\hat c}}_k^{e + 1}; \cdots ;{\bf{\hat c}}_k^E]$ is the set of intermediate representations of ${{\bf{C}}_k}$; ${\bf{c}}_k^i$$(e \le i \le E)$ is the representation of the $i$-th exit; ${{\bf{Z}}_{k - 1}}$ is equal to ${\bf{Z}}_{in}^{2k - 1}$ for adapter; ${{\bf{\bar C}}_{k - 1}} = [{\bf{c}}_{k - 1}^e;{\bf{c}}_{k - 1}^{e + 1}; \cdots ;{\bf{c}}_{k - 1}^E]$ (${{{\bf{\bar C}}}_{k - 1}} = {\bf{C}}_{in}^{2k - 1}$) is the set of exit-specific representations obtained by the ($k-1$)-th encoder layer. When a specific representation reaches its corresponding exit, it will end the forward propagation process. Therefore, if an exit is assigned before the $k$-th encoder layer, we have ${{\bf{C}}_{k - 1}} = [{\bf{c}}_{k - 1}^{e - 1};{\bf{c}}_{k - 1}^e;{\bf{c}}_{k - 1}^{e + 1}; \cdots ;{\bf{c}}_{k - 1}^E]$, otherwise, ${{\bf{\bar C}}_{k - 1}} = {{\bf{C}}_{k - 1}}$. For sample ${{\bf{X}}_j}$, we assume that the $e$-th exit is inserted after the  $\psi (e,{{\bf{X}}_j})$-th encoder layer. Accordingly, equation (6) is reformulated as
\begin{equation}
{\rm{O}} = \sum\limits_{e = 1}^E {\sum\limits_{j = 1}^N {\frac{1}{N}{\rm{C}}{{\rm{E}}_e}({\rm{Hea}}{{\rm{d}}_e}({\rm{LN}}({\bf{c}}_{\psi (e,{{\bf{X}}_j})}^e)),{y_j})} } ,
\label{eq16}
\end{equation}
where the representations for all exits are normalized by the same LN module which has been specifically trained for the prediction head in a pre-trained ViT.

\subsection{Graph Regularization}

In multi-exit neural networks, the consumed computational resources have a decisive influence on the performances of classifiers. The classifier at the deepest layer usually has the best performance because of the sufficiently computational resources \cite{b48,b49,b50}. Existing studies demonstrate that improving the learning abilities of classifiers at early exits is necessary to boost the inference efficiency of an early-exiting dynamic neural network \cite{b51,b52}. In MET, the exit-specific representations are fed into corresponding linear classifiers to make predictions. As for a linear classifier, its performance is normally affected by intra-class compactness and inter-class separability \cite{b53,b54,b55}. When the inputs exhibit poor intra-class compactness and small inter-class separability, it is hard to obtain a good linear classifier.

In this section, we boost the learning abilities of the classifiers with limited computational resources by improving the intra-class compactness and inter-class separability of their inputs. Graph has been proven to have the ability to effectively describe the relationship among samples \cite{b56,b57}. It includes vertices and edges, where vertices represent samples and edges are the corresponding similarities.  Similar to \cite{b58}, Euclidean distance is adopted to measure the difference between two data points. Accordingly, for the $e$-th classifier, the intra-class compactness of its inputs can be calculated by
\begin{equation}
{\rm{O}}_e^{{\rm{intra}}} = \sum\limits_{i = 1}^N {\sum\limits_{j = 1}^N {{{\left\| {{g_e}({{\bf{X}}_i},{{\bm{\theta }}_e}) - {g_e}({{\bf{X}}_j},{{\bm{\theta }}_e})} \right\|}_2}{\bf{S}}_{ij}^{{\rm{intra}}}} } ,
\label{eq17}
\end{equation}
where ${\left\|  \cdot  \right\|_2}$ is ${L_2}$ norm, ${\bf{S}}_{ij}^{{\rm{intra}}}$ is the intra-calss similarity between samples ${{\bf{X}}_i}$ and  ${{\bf{X}}_j}$. Similarly, the inter-class separability is formulated as
\begin{equation}
{\rm{O}}_e^{{\rm{inter}}} = \sum\limits_{i = 1}^N {\sum\limits_{j = 1}^N {{{\left\| {{g_e}({{\bf{X}}_i},{{\bm{\theta }}_e}) - {g_e}({{\bf{X}}_j},{{\bm{\theta }}_e})} \right\|}_2}{\bf{S}}_{ij}^{{\mathop{\rm inter}\nolimits} }} } ,
\label{eq18}
\end{equation}
where ${\bf{S}}_{ij}^{{\mathop{\rm inter}\nolimits} }$ is the inter-class similarity of the two samples.

Compared to the other classifiers, the classifier at the $E$-th exit has more advantages to capture the relationship among data points. As the performance of the $E$-th classifier is directly determined by its output logits, we construct the similarities ${\bf{S}}_{ij}^{{\rm{intra}}}$  and ${\bf{S}}_{ij}^{{\mathop{\rm inter}\nolimits} }$ as follows.
\begin{equation}
{\bf{S}}_{ij}^{{\rm{intra}}} = \left\{ \begin{array}{l}
{\rm{sim}}({\rm{logi}}{{\rm{t}}_i},{\rm{logi}}{{\rm{t}}_j}),{\rm{ }}(i,j) \in \pi ,\\
0,
\end{array} \right.,
\label{eq19}
\end{equation}
\begin{equation}
{\bf{S}}_{ij}^{{\mathop{\rm inter}\nolimits} } = \left\{ \begin{array}{l}
{\rm{sim}}({\rm{logi}}{{\rm{t}}_i},{\rm{logi}}{{\rm{t}}_j}),{\rm{ }}(i,j) \notin \pi ,\\
0,
\end{array} \right.,
\label{eq20}
\end{equation}
where ${\rm{logi}}{{\rm{t}}_i}$ is the output logit of the $E$-th classifier for ${{\bf{X}}_i}$; ${\rm{sim}}( \cdot , \cdot )$ stands for the similarity of two vectors; $(i,j) \in \pi $ means that samples ${{\bf{X}}_i}$ and ${{\bf{X}}_j}$ are from the same class while $(i,j) \notin \pi $  indicates that the two samples belong to different classes. For simplicity, cosine function is adopted to measure the similarity between samples.  Finally, the input representations for the $e$-th classifier can be enhanced by
\begin{equation}
\mathop {min}\limits_{{{\bm{\theta }}_e}} {\rm{ (}}\frac{1}{{{N^{{\rm{intra}}}}}}{\rm{O}}_e^{{\rm{intra}}} - \frac{1}{{{N^{{\rm{inter}}}}}}{\rm{O}}_e^{{\mathop{\rm inter}\nolimits} }),
\label{eq21}
\end{equation}
where ${N^{{\rm{intra}}}}$ and ${N^{{\rm{inter}}}}$ are the numbers of nonzero elements in ${\bf{S}}_{ij}^{{\rm{intra}}}$  and ${\bf{S}}_{ij}^{{\mathop{\rm inter}\nolimits} }$,  respectively.

\subsection{Objective Function}

As discussed above, cross-entropy losses are introduced to fine-tune a pre-trained ViT with multiple exits and graph regularized losses are designed to boost the performances of the classifiers at shallow layers. The total objective function of MET is formulated as
\begin{equation}
{\rm{Loss = O + }}\alpha \sum\limits_{e = 1}^{E - 1} {({\rm{O}}_e^{{\rm{intra}}} - {\rm{O}}_e^{{\mathop{\rm inter}\nolimits} })} ,
\label{eq22}
\end{equation}
where $\alpha $ is a trade-off parameter. Note that ${\bf{S}}_{ij}^{{\rm{intra}}}$ and ${\bf{S}}_{ij}^{{\mathop{\rm inter}\nolimits} }$ are constructed based on the output logits of the last classifier. As the logits should be optimized by the corresponding cross-entropy loss, we detach the gradients of the logits when constructing similarity matrices to avoid the negative influence brought by early exits.

\subsection{Discuss}

As indicated by equation (4), the representation of the class token in the $L$-th encoder layer is employed to make predictions in ViT. When ViT is assigned with multiple exits, we need to extract representations for the prediction heads at different exits. A simple approach is to allow these prediction heads to access the representations of the class token at corresponding depths. However, it is difficult for these classifiers to perform well because they may have conflicts in optimizing the shared representations of class token. The details are as follows.

Suppose the $e$-th and $(e+1)$-th exits are inserted after the $k$-th and $(k+1)$-th encoder layers. The cross-entropy loss of the $e$-th classifier is $\frac{1}{N}\sum\limits_{j = 1}^N {{\rm{C}}{{\rm{E}}_e}({\rm{Hea}}{{\rm{d}}_e}({\rm{LN}}({{\bf{c}}_k})),{y_j})} $. In the ($k+1$)-th encoder layer, the class token's representation ${{\bf{c}}_{k + 1}}$ is obtained by MHA and MLP blocks. In MHA, we have the following operations.
\begin{equation}
{\bf{c}}_{k + 1}^{at{t_i}} = sf(\frac{{({{{\bf{c'}}}_k}{\bf{W}}_{k + 1}^Q){{({{{\bf{B'}}}_k}{\bf{W}}_{k + 1}^K)}^{\rm{T}}}}}{{\sqrt {{d_h}} }})({{\bf{B'}}_k}{\bf{W}}_{k + 1}^V),
\label{eq23}
\end{equation}

\begin{equation}
{{\bf{\hat c}}_{k + 1}} = [{\bf{c}}_{k + 1}^{at{t_1}},{\bf{c}}_{k + 1}^{at{t_2}}, \cdots ,{\bf{c}}_{k + 1}^{at{t_h}}]{\bf{W}}_{k + 1}^{att} + {{\bf{c}}_k},
\label{eq24}
\end{equation}
where $sf( \cdot )$ represents the softmax function; $i = 1, \cdots ,h$; ${{\bf{B'}}_k} = [{{\bf{c'}}_k};{{\bf{Z'}}_k}]$; ${{\bf{c'}}_k}$ and ${{\bf{Z'}}_k}$ are the normalized ${{\bf{c}}_k}$ and ${{\bf{Z}}_k}$; ${\bf{W}}_{k + 1}^Q \in {\mathbb R^{d \times {d_h}}},{\bf{W}}_{k + 1}^K \in {\mathbb R^{d \times {d_h}}},{\bf{W}}_{k + 1}^V \in {\mathbb R^{d \times {d_h}}}$ are applied to obtain query, key and value matrices; ${d_h} = \frac{d}{h}$ and $h$ is the number of attention heads; ${\bf{c}}_{k + 1}^{at{t_i}}$ is the output of the $i$-th attention head; ${\bf{W}}_{k + 1}^{att} \in {\mathbb R^{d \times d}}$ is a mapping matrix. In MLP, we have
\begin{equation}
{{\bf{c}}_{k + 1}} = {\rm{GELU(}}{{\bf{\hat c'}}_{k + 1}}{\bf{W}}_{k + 1}^{up}){\bf{W}}_{k + 1}^{down} + {{\bf{\hat c}}_{k + 1}},
\label{eq25}
\end{equation}
where ${\rm{GELU}}( \cdot )$ represents the GELU function; ${\bf{W}}_{k + 1}^{up} \in {\mathbb R^{d \times 4d}}$ and ${\bf{W}}_{k + 1}^{down} \in {\mathbb R^{4d \times d}}$ are up-projection and down-rpojection matrices; ${{\bf{\hat c'}}_{k + 1}}$ is the normalized ${{\bf{\hat c}}_{k + 1}}$. At the $(e+1)$-th exit, the cross-entropy loss can be formulated as  $\frac{1}{N}\sum\limits_{j = 1}^N {{\rm{C}}{{\rm{E}}_{e + 1}}({\rm{Hea}}{{\rm{d}}_{e + 1}}({\rm{LN}}({{\bf{c}}_{k + 1}})),{y_j})} $.

For an early-exiting dynamic neural network, it is optimized by jointly minimizing the losses of all exits. When directly attaching classifiers along the network's depth, classifiers at early exits exert a detrimental effect on the subsequent representation-learning processes \cite{b50}. According to equations (23)-(25), ${{\bf{c}}_{k + 1}}$ heavily depends on the class token's representation obtained from the $k$-th encoder layer. During the joint optimization process, the loss at the $e$-th exit probably hurts the quality of ${{\bf{c}}_{k + 1}}$, negatively affecting the performance of the $(e+1)$-th classifier.

\section{Experiments}

In this section, extensive experiments are conducted to evaluate the performance of MET. Specifically, we first verify the adaptation ability of MET on VTAB-1K benchmark \cite{b59}, then test its generalization ability on few-shot learning and domain generalization benchmarks, and finally conduct ablation studies for further analysis of the proposed method. In this paper, we select the ViT-B/16 \cite{b2} pre-trained on supervised ImageNet-21K \cite{b7} as the backbone for all compared methods and MET. In MET, we assign 7 prediction heads to the ViT-B/16 backbone. To be specific, for each of the last 7 encoder layers, it is followed by a prediction head. In other words, the $e$-th prediction head is inserted after the ($L-7+e$)-th encoder layer. For compared methods, we directly run the released code or use the reported results. Following other dynamic neural networks \cite{b19,b20,b48,b49,b50,b51}, the computational budget of each method is measured in GFLOPs.

\begin{table*}[!htb]
  \centering
  \caption{Experimental results on VTAB-1K Benchmark. 'Group Mean' represents the average accuracy of the three groups. 'Param. (M)' denotes the number of trainable parameters and '\#Param. (M)' means the corresponding group-wise value. 'GFLOPs' represents the average GFLOPs across all datasets. In parameter-efficient tuning methods, \textbf{bold font} and \underline{underline} denote the \textbf{best} and the \underline{second-best} results. }
  \label{tab1}
  \renewcommand{\arraystretch}{1.2}
  \setlength\tabcolsep{3pt}

  \begin{tabular}{l|ccccccc|cccc|cccccccc|ccc}
    \toprule
    &\multicolumn{7}{c|}{{\textbf{Natural}}}  &\multicolumn{4}{c|}{{\textbf{Specialized}}}  &\multicolumn{8}{c|}{{\textbf{Structured}}}
    &

    \\
       \textbf{Method}        &\rotatebox{90}{Cifar100}   &\rotatebox{90}{Caltech101}     &\rotatebox{90}{DTD}    &\rotatebox{90}{Flower102}   &\rotatebox{90}{Pets}   &\rotatebox{90}{SVHN}   &\rotatebox{90}{Sun397} &\rotatebox{90}{Camelyon}&\rotatebox{90}{EuroSAT}&\rotatebox{90}{Resisc45}&\rotatebox{90}{Retinopathy}&\rotatebox{90}{Clevr-Count}&\rotatebox{90}{Clevr-Dist}&\rotatebox{90}{DMLab}&\rotatebox{90}{KITTI-Dist}&\rotatebox{90}{dSpr-Loc}&\rotatebox{90}{dSpr-Ori}&\rotatebox{90}{sNORB-Azim}&\rotatebox{90}{sNORB-Ele}&\rotatebox{90}{\textbf{Group Mean}}&\rotatebox{90}{\textbf{\#Param. (M)}}&\rotatebox{90}{\textbf{GFLOPs}}\\

              \hline
     \multicolumn{23}{c}{\textit{Traditional fine-tuning (static inference)}}   \\\hline

    Full  &68.9 &87.7 &64.3 &97.2 &86.9 &87.4 &38.8 &79.7 &95.7 &84.2 &73.9 &56.3 &58.6 &41.7 &65.5 &57.5 &46.7 &25.7 &29.1  &69.0 &85.8 &17.58
      \\
    Linear &63.4 &85.0 &63.2 &97.0 &86.3 &36.6 &51.0 &78.5 &87.5 &68.6 &74.0 &34.3 &30.6 &33.2 &55.4 &12.5 &20.0 &9.6 &19.2 &57.6 &0.04 &17.58        \\\hline
    \multicolumn{23}{c}{\textit{Parameter-efficient tuning methods (static inference)}}   \\\hline
     Adapter \cite{b29} &69.2 &90.1 &68.0 &98.8 &89.9 &82.8 &54.3 &84.0 &94.9 &81.9 &75.5 &80.9 &65.3 &48.6 &78.3 &74.8 &48.5 &29.9 &41.6 &73.9 &0.20 &17.61
    \\
    LoRA \cite{b43} &67.1 &91.4 &69.4 &98.8 &90.4 &85.3 &54.0 &84.9 &95.3 &84.4 &73.6 &
\underline{82.9} &69.2 &49.8 &78.5 &75.7 &47.1 &31.0 &44.0 &74.6 &0.33 &
17.58
    \\
    NOAH \cite{b45} &69.6 &
92.7 &70.2 &99.1 &90.4 &86.1 &53.7 &84.4 &95.4 &83.9 &75.8 &82.8 &68.9 &49.9 &81.7 &81.8 &48.3 &32.8 &44.2 &75.5 &0.40 &
17.58
    \\

    VPT \cite{b17}  &\textbf{78.8} &90.8 &65.8 &98.0 &88.3 &78.1 &49.6  &81.8 &
96.1 &83.4 &68.4 &68.5 &60.0 &46.5 &72.8 &73.6 &47.9 &32.9 &37.8 &72.0 &0.57 &18.30 \\

    FacT-TT \cite{b41} &71.3 &89.6 &70.7 &98.9 &91.0 &87.8 &54.6 &85.2 &95.5 &83.4 &75.7 &82.0 &69.0 &49.8 &80.0 &79.2 &48.4 &
34.2 &41.4 &75.3 &\textbf{0.08} &
17.58
    \\

    ARC \cite{b13} &72.2 &90.1 &
\underline{72.7} &99.0 &91.0 &\textbf{91.9} &54.4 &84.9 &95.7 &\textbf{86.7} &75.8 &80.7 &67.1 &48.7 &81.6 &79.2 &51.0 &31.4 &39.9 &75.8 &
\underline{0.13} &
17.58
    \\
    Res-Tuning \cite{b15} &75.2 &92.7 &71.9 &\textbf{99.3} &\textbf{91.9} &86.7 &\textbf{58.5} &86.7 &95.6 &85.0 &74.6 &80.2 &63.6 &50.6 &80.2 &\textbf{85.4} &\textbf{55.7} &31.9 &42.0 &76.3 &0.55 &17.67
    \\
    Hydra \cite{b40} &72.7 &91.3 &72.0 &
\underline{99.2} &91.4 &90.7 &55.5  &85.8 &96.0 &86.1 &75.9 &\textbf{83.2} &68.2 &50.9 &
82.3 &80.3 &50.8 &\underline{34.5} &43.1 &76.5 &0.28 &
17.58
   \\
   RLRR \cite{b42}&
\underline{75.6} &92.4 &\textbf{72.9} &\textbf{99.3} &
91.5 &89.8 &
\underline{57.0} &
86.8 &95.2 &85.3 &75.9 &79.7 &64.2 &\textbf{53.9} &82.1 &
\underline{83.9} &53.7 &33.4 &43.6 &76.8 &0.33&
17.58
   \\
   SCT \cite{b14} &75.3 &91.6 &72.2 &
\underline{99.2} &91.1 &
\underline{91.2} &55.0 &85.0 &
96.1 &
\underline{86.3} &
76.2 &81.5 &65.1 &
\underline{51.7} &80.2 &75.4 &46.2 &33.2 &
45.7 &76.0 &0.15&17.60
   \\

         \rowcolor{gray!20}
   MET(static) &72.5 &94.3 &71.2 &\textbf{99.3} &90.9 &88.0 &56.4 &\underline{88.8} &\underline{96.4} &85.9 &\underline{76.3} &82.1 &
69.1 &49.3 &\underline{83.4} &\textbf{85.4} &
54.7 &33.5 &\underline{49.3} &\underline{77.3} &
\underline{0.13}&15.61
   \\
   \rowcolor{gray!20}
   (Exit) &\textit{7} &\textit{5} &\textit{6} &\textit{6} &\textit{6} &\textit{3} &\textit{7} &\textit{1} &\textit{4} &\textit{5} &\textit{3} &\textit{6} &\textit{3} &\textit{4} &\textit{4} &\textit{5} &\textit{4} &\textit{5} &\textit{6} &\textendash &\textendash&\textendash
   \\
   \hline
     \multicolumn{23}{c}{\textit{Parameter-efficient tuning methods (dynamic inference)}}  \\\hline

   DyT($r = 0.5$)\cite{b46} &70.4 &94.2 &71.1 &99.1 &\underline{91.7} &88.0 &51.5 &87.1 &95.3 &84.2 &75.8 &79.2 &61.8 &51.0 &82.4 &79.7 &52.3 &\textbf{35.3} &44.5 &75.7 &0.20 &12.54
   \\
    \rowcolor{gray!20}
    MET(small) &70.4 &94.2 &71.3 &99.1 &91.4 &87.2 &51.8 &\underline{88.8} &96.0 &84.5 &76.1 &80.4&\underline{69.5}&50.3&81.0&83.6&54.1&30.4&46.6                                       &76.5 &0.20  &\textbf{12.47}
   \\\hline

   DyT($r = 0.9$)\cite{b46} &74.0 &\textbf{95.1} &\textbf{72.9} &\textbf{99.3} &\underline{91.7} &87.6 &56.9 &87.7 &95.7 &85.4 &76.1 &81.6 &63.2 &50.1 &83.0 &83.3 &52.0 &\underline{34.5} &44.5 &76.7 &0.20 &17.07
   \\
   \rowcolor{gray!20}
       MET(large) &72.5 &\underline{94.4} &71.7&\textbf{99.3} &91.6 &88.2 &56.4 &\textbf{89.2} &\textbf{96.5} &86.1 &\textbf{76.6} &82.5 &\textbf{70.0} &50.3 &\textbf{83.5} &\textbf{85.4} &\underline{54.9} &33.9 &\textbf{50.2}
        &\textbf{77.7} &0.20  &\underline{13.99}
   \\

\bottomrule
  \end{tabular}
\end{table*}

\begin{table}[!htb]
  \centering
  \caption{Details of VTAB-1K Benchmark.}
  \label{tab2}
    \renewcommand{\arraystretch}{1.2}
  \setlength\tabcolsep{3pt}

  \begin{tabular}{lccccc}
    \toprule
    \textbf{Dataset}      &\textbf{Description}   &\#\textbf{Class}   &\#\textbf{Train}  &\#\textbf{Val}   &\#\textbf{Test}  \\ \hline

   Cifar100    &\multirow{7}{*}{Natural}     &100      &\multirow{7}{*}{800/1000}     &\multirow{7}{*}{200}  &10000     \\

   Caltech101    &                            &102     &  &        &6084     \\

   DTD          &                             &47      &   &         &1880    \\
   Flower102    &                             &102     &  &           &6149    \\

    Pets       &                             &37       &   &        & 3669   \\
    SVHN       &                             &10       &  &         &26032    \\
    Sun397    &                              &397      &  &          &21750    \\
       \hline

  Camelyon         &\multirow{4}{*}{Specialized}      &2     &\multirow{4}{*}{800/1000}   &\multirow{4}{*}{200}   &32768 \\

   EuroSAT           &           &10        &     &    &5400 \\
   Resisc45          &           &45        &     &    &6300 \\
   Retinopathy       &           &5        &     &    &42670 \\
      \hline
  Clevr-Count         &\multirow{8}{*}{Structured}            &8         &\multirow{8}{*}{800/1000}   &\multirow{8}{*}{200}     &15000     \\
  Clevr-Dist         &           &6        &     &   &15000  \\
  DMLab              &           &6        &     &    &22735 \\
  KITTI-Dist         &           &4        &     &    &711 \\
  dSpr-Loc           &           &16        &     &    &73728 \\
  dSpr-Ori           &           &16        &     &    &73728 \\
  sNORB-Azim         &           &18        &     &    &12150 \\
  sNORB-Ele          &           &9        &     &    &12150 \\

\bottomrule
  \end{tabular}
\end{table}

\begin{table}[!htb]
  \centering
  \caption{Parameter settings of MET on VTAB-1K Benchmark.}
  \label{tab3}
    \renewcommand{\arraystretch}{1.2}
  \setlength\tabcolsep{2pt}
  \begin{tabular}{lc}
    \toprule
    \textbf{Name}      &\textbf{Value}  \\ \hline

    Optimizer          &Adam                           \\
Learning rate          &\{0.03,0.01, 0.005,0.003,0.001, 0.0005\}\\
Weight decay           &\{0.05, 0.01, 0.005, 0.001, 0.0\}\\
Batch size             &32  \\
Learning rate schedule	&Cosine Decay\\
Warmup epochs	&10\\
Training epochs	&100\\
$\alpha $	&\{0.1,0.01,0.001\}\\
${d'}$	 &\{30,50\}\\

\bottomrule
  \end{tabular}
\end{table}

\subsection{Experiments on VTAB-1K Benchmark}

The comparison results on VTAB-1K benchmark are reported in TABLE~\ref{tab1}. VTAB-1K is a widely used transfer learning benchmark. It contains 19 visual classification datasets which can be divided into three categories: \textbf{Natural}, \textbf{Specialized}, and \textbf{Structured}. Each dataset only has 1000 training data points. The details are given in TABLE~\ref{tab2}, where '\#Class' represents the number of classes, '\#Train', '\#Val' and '\#Test' denote the number of samples in training, validation and test sets, respectively. Similar to VPT \cite{b17}, we apply standard data augmentations, including image normalization and resizing each input into a $224 \times 224$ image, for these samples.

\begin{figure*}[bp]
\centering
\includegraphics[trim={0.2cm 0.cm 0cm 0.4cm},clip,width=7.2in]{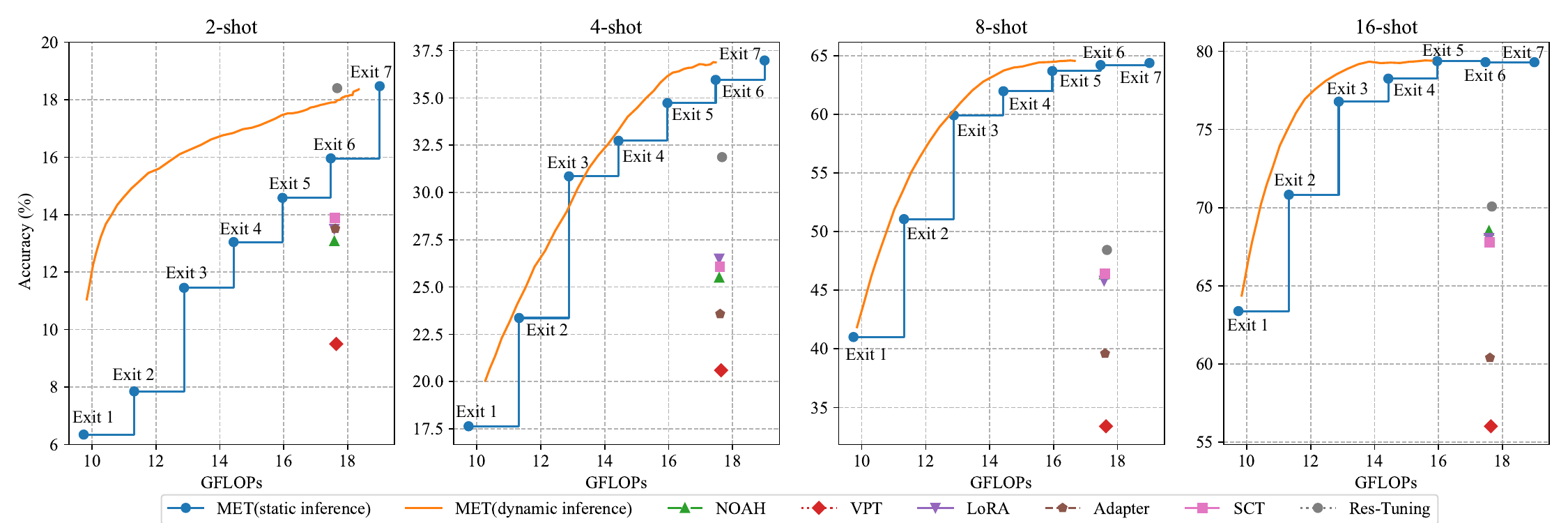}
\caption{Experimental results on Stanford Cars dataset.}
\label{fig5}
\end{figure*}
\begin{figure*}[bp]
\centering
\includegraphics[trim={0.2cm 0.0cm 0cm 0.4cm},clip,width=7.2in]{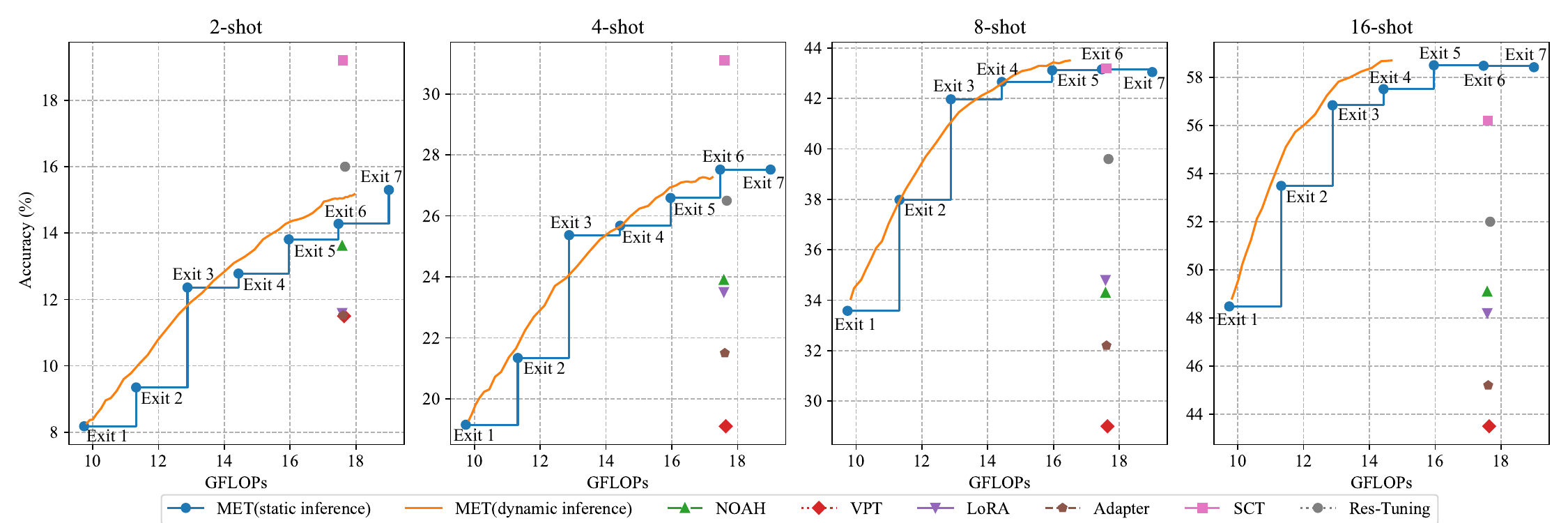}
\caption{Experimental results on FGVC-Aircraft dataset.}
\label{fig6}
\end{figure*}

To evaluate the performance of MET, we compare it with two traditional baselines and several competitive PETL methods. The two traditional baselines are full fine-tuning (denoted as Full) which updates the whole pre-trained backbone, and linear probing (denoted as Linear) which only updates the prediction head. PETL methods include Adapter \cite{b29}, LoRA \cite{b43}, NOAH \cite{b45}, VPT \cite{b17}, FacT-TT \cite{b41}, ARC \cite{b13}, Res-Tuning \cite{b15}, Hydra \cite{b40}, RLRR \cite{b42}, SCT \cite{b14}, and DyT \cite{b46}. The parameter settings of MET are presented in TABLE~\ref{tab3}. Following \cite{b13,b14,b15,b17}, we also use the validation set split from the training set to determine the hyper-parameters in TABLE~\ref{tab3} and then directly report the top-1 accuracy on the test set for each dataset. As a dynamic neural network, MET has two prediction modes: anytime classification (static inference) and budgeted batch classification (dynamic inference) \cite{b48,b49,b50,b51}. In static inference mode, we only need to reserve a suitable prediction head and store the extra parameters inserted into the backbone. In the other inference mode, to ensure a fair comparison with DyT, we let MET have a comparable parameter scale. To achieve this goal, we only reserve three prediction heads. In this case, MET is run with prediction heads \{1,4,7\} or \{5,6,7\}. For a better comparison, we denote the dynamic MET with smaller/larger GFLOPs as MET(small)/MET(large).

Note that in static inference mode, MET only has one prediction head, and thus 'Exit' in TABLE~\ref{tab1} refers to the corresponding exit. From TABLE~\ref{tab1}, we have the following observations. 1) Although PETL methods with static inference obviously perform better than linear probing and full fine-tuning, most of them have the same computational cost as the two traditional tuning methods, and even some of them, such as Adapter, VPT, Res-Tuning, and SCT, introduce extra GFLOPs. 2) In static inference mode, MET outperforms all compared methods in terms of classification accuracy and computational cost. Specifically, it exceeds the method with the second-best performance (RLRR) by 0.5\% and simultaneously reduces GFLOPs by 11.2\%. 3) In dynamic inference mode, compared to DyT, MET can achieve more competitive classification performance with significantly fewer GFLOPs, while maintaining a comparable parameter scale. These observations demonstrate that our proposed method is both storage-friendly and inference-efficient.

\begin{figure*}[!htb]
\centering
\includegraphics[trim={0.1cm 0cm 0cm 0.4cm},clip,width=7.2in]{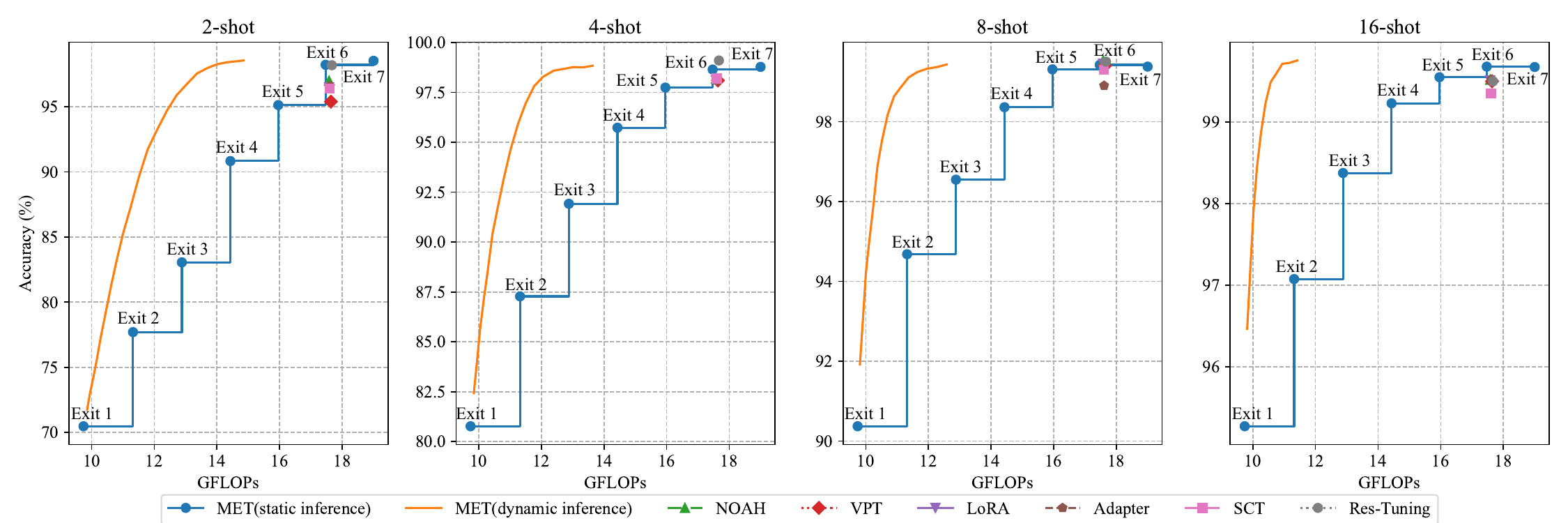}
\caption{Experimental results on Oxford-Flowers102 dataset.}
\label{fig7}
\end{figure*}
\begin{figure*}[!htb]
\centering
\includegraphics[trim={0.1cm 0cm 0cm 0.4cm},clip,width=7.2in]{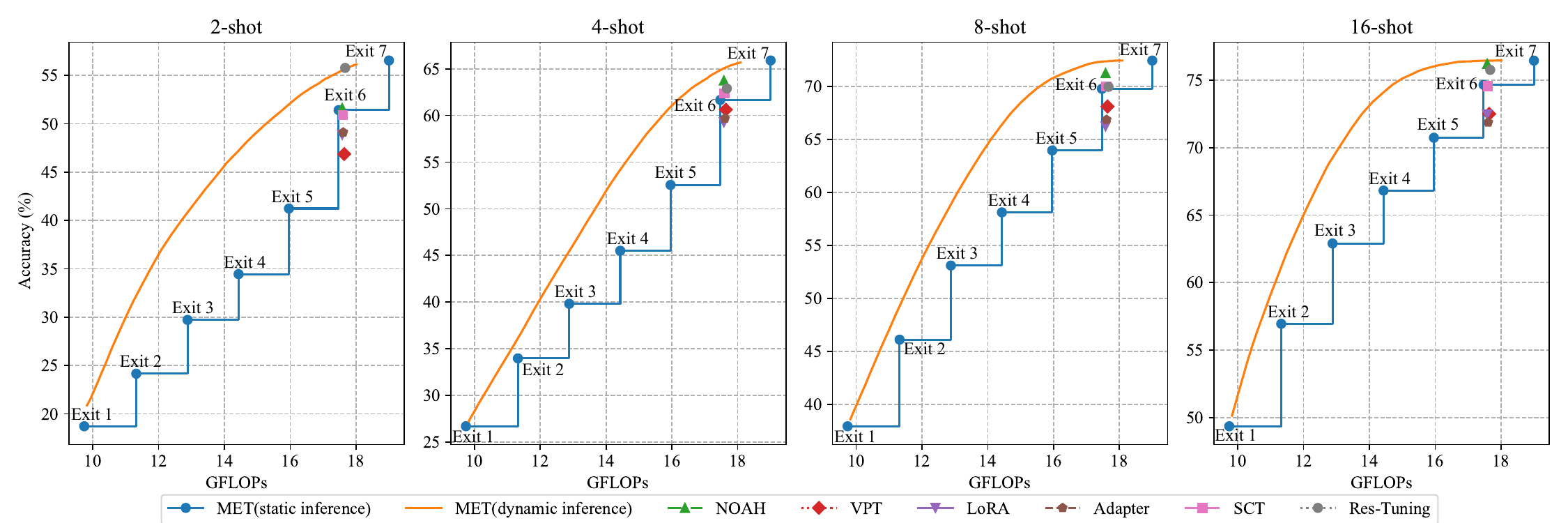}
\caption{Experimental results on Food101 dataset.}
\label{fig8}
\end{figure*}
\begin{figure*}[!htb]
\centering
\includegraphics[trim={0.1cm 0cm 0cm 0.1cm},clip,width=7.2in]{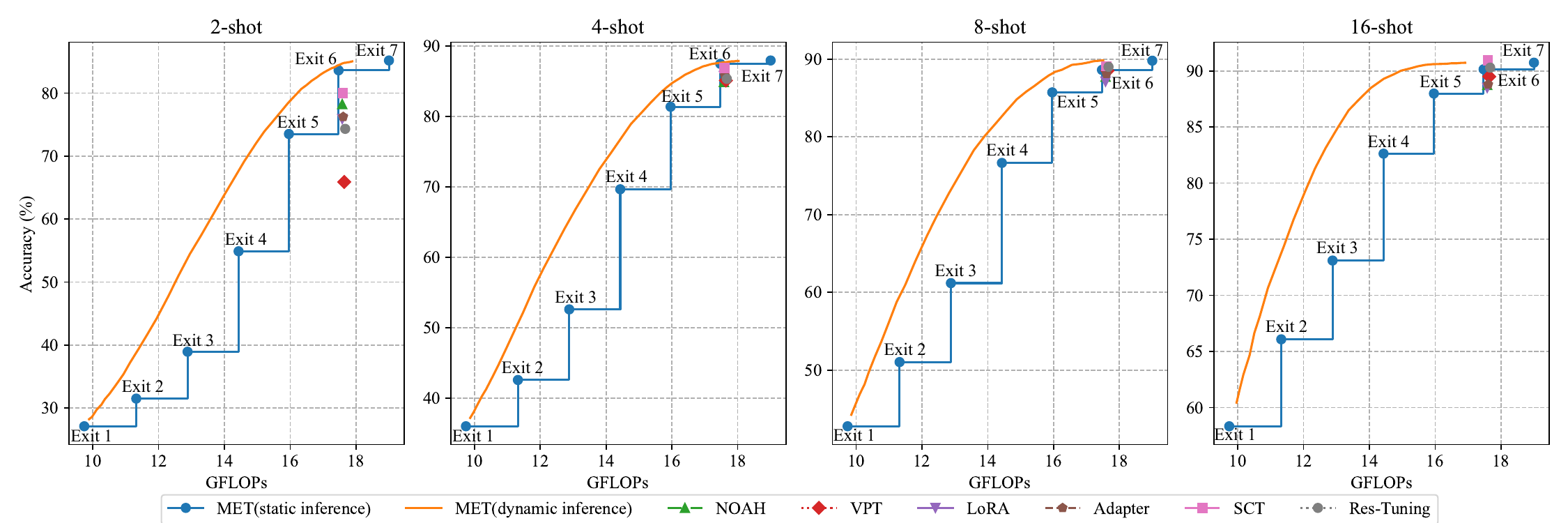}
\caption{Experimental results on Oxford-Pets dataset.}
\label{fig9}
\end{figure*}

\begin{figure*}[!htb]
\centering
\includegraphics[trim={0.1cm 0cm 0cm 0.1cm},clip,width=7.2in]{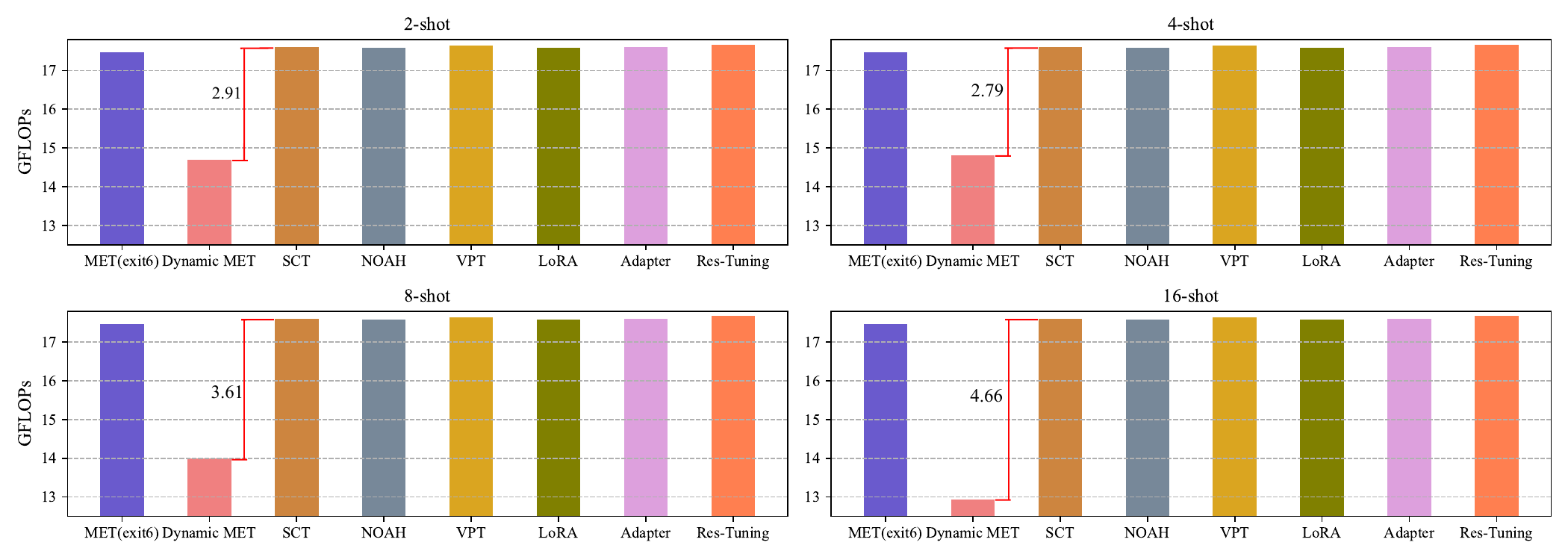}
\caption{Average GFLOPs on fine-grained datasets.}
\label{fig10}
\end{figure*}

\subsection{Experiments on Few-Shot Learning}

Similar to \cite{b14,b15,b45}, we select five fine-grained visual recognition datasets in few-shot settings to test the learning ability of our proposed method under low-data regime. The details of these datasets are described in TABLE~\ref{tab4}. We evaluate MET under \{2,4,8,16\}-shot settings. TABLE~\ref{tab5} shows the corresponding parameter-settings for MET. We choose Adapter \cite{b29}, LoRA \cite{b43}, NOAH \cite{b45}, VPT \cite{b17}, Res-Tuning \cite{b15}, and SCT \cite{b14} as comparison methods, as their advantages in few-shot learning have been verified by extensive experiments.

\begin{figure}[!h]
\centering
\includegraphics[trim={0.3cm 0.5cm 0.0cm 0.52cm},clip,width=3.2in]{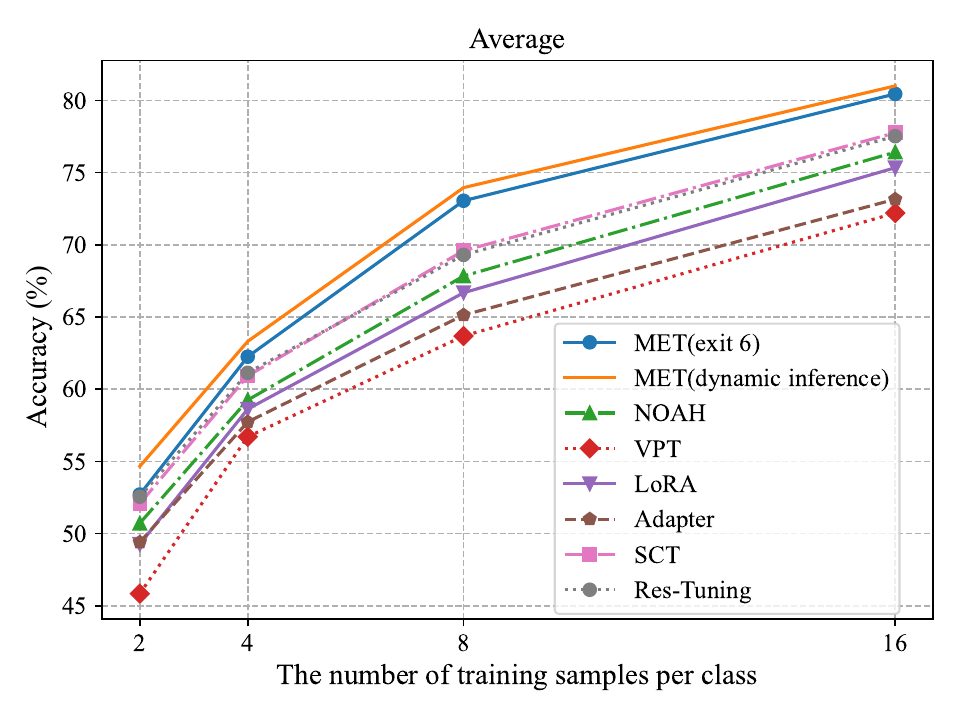}
\caption{Average accuracies on fine-grained datasets.}
\label{fig11}
\end{figure}

\begin{table}[!htb]
  \centering
  \caption{Details of five fine-grained datasets.}
  \label{tab4}
      \renewcommand{\arraystretch}{1.2}
  \setlength\tabcolsep{3pt}
  \begin{tabular}{lcccc}
    \toprule
    \textbf{Dataset}         &\#\textbf{Class}   &\#\textbf{Train}  &\#\textbf{Val}   &\#\textbf{Test}  \\ \hline

   Stanford Cars      &196      &\multirow{5}{*}{(2/4/8/16)$\times$\#Class}    &1635  &8041    \\

   FGVC-Aircraft        &100     &  &3333        &3333     \\
   Oxford-Flowers102   &102      &  &1633 &2463\\
   Food101             &101      &  &20200 &30300\\
   Oxford-Pets         &37      &   &736&3669\\

\bottomrule
  \end{tabular}
\end{table}

\begin{table}[!htb]
  \centering
  \caption{Parameter settings of MET on few-shot learning.}
  \label{tab5}
      \renewcommand{\arraystretch}{1.2}
  \setlength\tabcolsep{3pt}
  \begin{tabular}{lc}
    \toprule
    \textbf{Name}      &\textbf{Value}  \\ \hline

    Optimizer          &Adam                           \\
Learning rate          &\{0.03,0.01, 0.005,0.003,0.001, 0.0005\}\\
Weight decay           &\{0.05, 0.01, 0.005, 0.001, 0.0\}\\
Batch size             &32  \\
Learning rate schedule	&Cosine Decay\\
Warmup epochs	&10\\
Training epochs	&100\\
$\alpha $	&0.01\\
${d'}$	 &30\\

\bottomrule
  \end{tabular}
\end{table}

Figs.~\ref{fig5}-\ref{fig9} show experimental results on fine-grained datasets. On the datasets of Stanford Cars, Food101, and Oxford-Pets, dynamic MET evidently outperforms the compared methods in terms of GFLOPs and accuracy. On the Oxford-Flowers102 dataset with \{2,4,8,16\}-shot settings, dynamic MET achieves the similar accuracies of the second-best method while reducing GFLOPs by 20.7\%, 22.8\%, 28.8\% and 40.2\%, respectively. On FGVC-Aircraft dataset, both static MET and dynamic MET underperform SCT in 2-shot and 4-shot settings but they are more competitive than SCT in the other settings. From these figures, it is easy to find that static MET with the 6-th exit performs better than many of these compared methods in most cases while consuming fewer GFLOPs. Here, we show the average GFLOPs and the overall performance in Figs.~\ref{fig10} and~\ref{fig11}. According to the results in Fig.~\ref{fig11}, we can observe that MET has an obvious advantage over Adapter, LoRA, NOAH, VPT, Res-Tuning and SCT in two inference modes. Fig.~\ref{fig11} displays that SCT and Res-Tuning perform better than the other compared methods and there is little difference in learning performance between the two methods. Taking SCT as an example, we record the GFLOPs required for dynamic MET to achieve the classification performance similar to that of SCT. Note that we record the GFLOPs of the best performance if dynamic MET underperforms SCT. The results presented in Fig.~\ref{fig10} clearly illustrate that dynamic MET requires the fewest GFLOPs. Compared to SCT, dynamic MET can reduce GFLOPs by 16.5\%, 15.8\%, 20.5\%, and 26.5\% in \{2,4,8,16\}-shot settings. These phenomena demonstrate the efficiency and effectiveness of MET under low-data regime.

\begin{figure*}[!htb]
\centering
\includegraphics[trim={0.1cm 0cm 0cm 0.1cm},clip,width=7.2in]{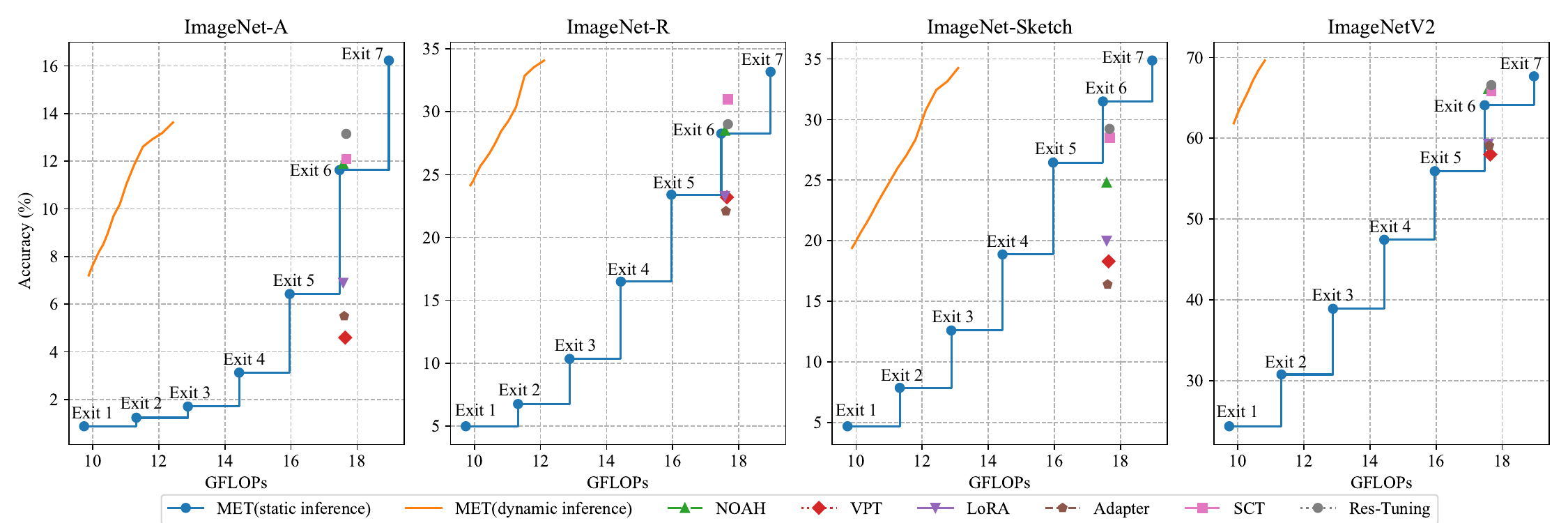}
\caption{Experimental results on domain generalization benchmarks.}
\label{fig12}
\end{figure*}
\begin{table*}[!htb]
  \centering
  \caption{Ablation experiments on VTAB-1K benchmark.}
  \label{tab6}
     \renewcommand{\arraystretch}{1.2}
  \setlength\tabcolsep{3pt}
  \begin{tabular}{lccccccccccc}

    \toprule
    \textbf{Notation}    &\textbf{Before MHA}   &\textbf{Before FFN}  &\#\textbf{GR}   &\textbf{Exit 1} &\textbf{Exit 2} &\textbf{Exit 3} &\textbf{Exit 4} &\textbf{Exit 5}  &\textbf{Exit 6} &\textbf{Exit 7} &\textbf{Average}\\ \hline

   METMG      &$\checkmark$  &$\times$       &$\checkmark$    &69.30 &71.13	&72.71	&73.90	&75.03	&75.67	&76.11	&73.41  \\

   METFG       &$\times$     &$\checkmark$    &$\checkmark$   &68.95 &71.07	&72.67	&73.96	&75.21	&75.98	&76.35	&73.46     \\
   METMF       &$\checkmark$    &$\checkmark$ &$\times$    &69.35	&71.33	&71.39	&74.37	&75.57	&76.18	&76.43	&73.52\\
   MET         &$\checkmark$    &$\checkmark$ &$\checkmark$    &\textbf{70.33}	&\textbf{72.26}	&\textbf{73.85}	&\textbf{75.20}	&\textbf{76.38}	&\textbf{77.11}	&\textbf{77.35}	&\textbf{74.64}\\

\bottomrule
  \end{tabular}
\end{table*}

\subsection{Experiments on Domain Generalization}

In some real-world applications, the distribution of test data is different from that of training data. The trained models usually suffer from performance drops in this circumstance. To evaluate the generalization ability of MET to domain shift, following \cite{b14,b15,b45}, we train our method on ImageNet dataset and directly test it on four variants of this dataset. To be specific, the training set sampled from ImageNet-1K contains 16 images per category and test sets include ImageNet-A \cite{b60}, ImageNet-R \cite{b61}, ImageNet-Sketch \cite{b62}, and ImageNetV2 \cite{b63}. ImageNet-A and ImageNet-R contain adversarially-filtered examples and artistic renditions of ImageNet-1K. ImageNet-Sketch consists of the sketch images of the same categories in ImageNet. Compared to ImageNet, ImageNetV2 is collected from different sources. For the purpose of comparison, we also report the experimental results of Adapter, LoRA, NOAH, VPT, Res-Tuning and SCT on test sets. For MET, we follow the parameter settings described in TABLE~\ref{tab5}.

Experimental results are reported in Fig.~\ref{fig12}. The static MET at the 6-th exit requires less computational cost than Adapter, LoRA, NOAH, VPT, Res-Tuning, and SCT. At this exit, we find that: 1) static MET has a clear advantage over VPT, LoRA, and Adapter, and performs better than all compared methods on ImageNet-Sketch dataset; 2) the performance of static MET is similar to that of NOAH. In dynamic inference mode, MET significantly outperforms all compared methods on these test sets. When achieving the performance similar to that of the second-best compared method on ImageNet-A, ImageNet-R, ImageNet-Sketch, and ImageNetV2 datasets, dynamic MET can reduce GFLOPs by 31.5\%, 34.8\%, 31.5\%, and 40.8\%. The above comparisons illustrate MET's superiority in handling the domain shift problem, thereby demonstrating its robust generalization ability.

\subsection{Ablation Studies}

MET consists of E-adapters and graph regularization. As demonstrated in Section IV.A, E-adapters are inserted before the MHA and FFN modules in each encoder layer. Here, we investigate the influence of different components on the performance of MET. Without loss of generality, we conduct experiments on VTAB-1K benchmark.

\begin{figure}[!h]
\centering
\includegraphics[trim={0.2cm 0cm 0cm 0.1cm},clip,width=3.4in]{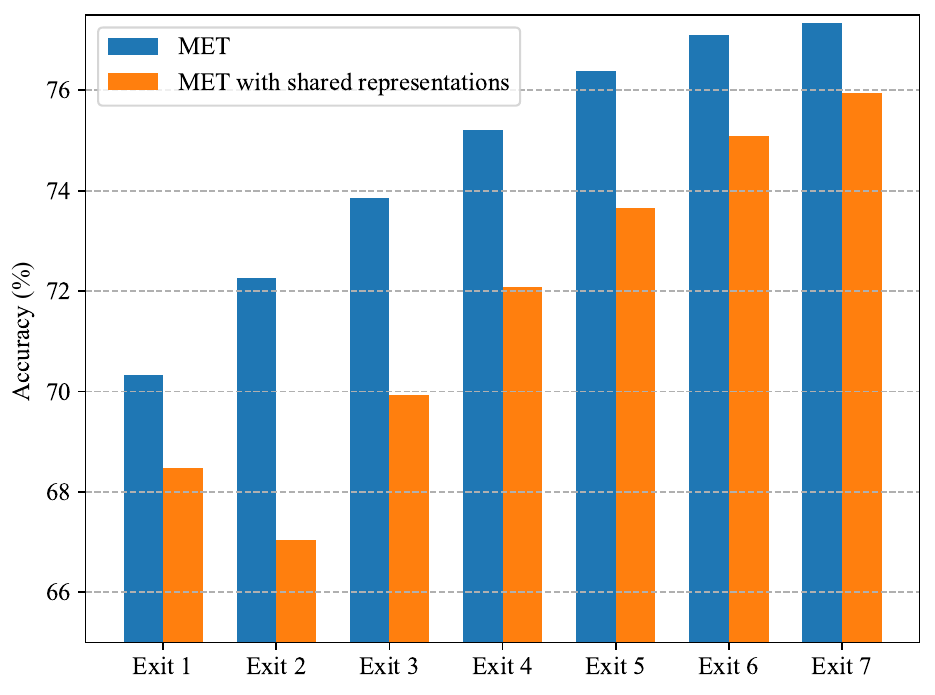}
\caption{Comparison between exit-specific information extraction and the one with shared representations.}
\label{fig13}
\end{figure}

To illustrate the impact on different exits, we report the group-wise average accuracies of all exits in TABLE~\ref{tab6}, where '\#GR' represents graph regularization and the best results are highlighted in bold. For convenience, we give the notations in the first column. For example, in METMG, E-adapters are inserted before MHA, and early exits are supervised by graph regularization. The performance of METMG is similar to that of METFG, indicating that the two locations are equally important. In comparison with METMF, MET shows a clear advantage across all exits and can obtain a 1.12\% gain in terms of average accuracy. This observation demonstrates that graph regularization has the ability to enhance the performances of early classifiers. Comparing the experimental results of METMG, METFG and METMF, we find that the three factors contribute similarly to the overall performance. All in all, the E-adapters from the two locations  and graph regularization are essential for MET.

In the proposed method, E-adapters are introduced to extract exit-specific representations. As analyzed in section IV.D, if multiple exits share the representations of class token, some optimization conflicts may negatively affect the final performance. Here, we only keep one diagonal
matrix in MET to demonstrate the influence. As displayed in Fig.~\ref{fig13}, MET with shared representations performs much worse than MET. Since different exits have different requirements for the intermediate representations of class token in encoder layers, these requirements probably conflict with each other. The exit at deeper depth usually has a better performance because of more computational resources. However, the second exit underperforms the first one in MET with shared representations, illustrating that the loss at the second exit is not well-optimized. According to these results, our proposed method can effectively solve the optimization conflicts.

\section{Conclusion and future work}

In this paper, a novel inference-efficient transfer learning method, named multiple-exit tuning (MET), is introduced to adapt a ViT pre-trained on large-scale datasets to various downstream domains. Unlike most existing PETL methods, MET focuses on assigning suitable computational resources for data points during inference stage. To achieve this goal, multiple exits are inserted into a pre-trained ViT backbone in MET, which allows samples to flow to suitable exits, thereby reducing the computational cost of processing easy samples. MET is composed of E-adapters and graph regularization. E-adapters are employed to extract appropriate representations for different exits. To reduce the number of trainable parameters, E-adapters are designed with a partial-sharing projection scheme. The classifiers at early exits usually underperform the one at the deepest exit due to limited computational resources. Two supervised graphs, which characterize the intra-class and inter-class relationship among samples, are adopted to regularize the representation-learning processes of early exits. We evaluate the performance of MET on 28 downstream tasks. As demonstrated by experimental results, MET achieves the state-of-the-art performance in terms of both classification accuracy and inference efficiency.

Early-exiting dynamic neural networks usually have difficulties in determining whether an input is easy or not at the image level for some tasks such as object detection and semantic segmentation \cite{b48,b49,b50,b51,b52}. This inherent characteristic implies that our proposed method is mainly proposed for image classification tasks. As for future work, we plan to explore effective strategies to expand the application scope of MET.


%


\ifCLASSOPTIONcaptionsoff
  \newpage
\fi

\end{document}